\documentclass[preprint,12pt,sort&compress,review]{elsarticle}

\usepackage{amssymb,graphics,subfigure}
\usepackage{amsmath}
\usepackage{amsfonts}  
\usepackage{floatflt}
\usepackage{indentfirst}
\usepackage{latexsym,bm,amssymb}
\usepackage{multicol,multirow,booktabs}
\usepackage{rotating}

\journal{Journal of Electronic Imaging}

\begin{document}

\begin{frontmatter}

\title{Effective scaling registration approach by imposing the emphasis on the scale factor}


\author{}
\address{Minmin Xu$^{1}$, Siyu Xu$^{1}$, Jihua Zhu$^{*1}$, Yaochen Li$^{1}$, Jun Wang$^{2}$, Huimin Lu$^{3}$}
\address{1. School of Software Engineering, Xi'an Jiaotong University, Xi'an 710049, China\\
         2. School of Digital Media, Jiangnan University, Wuxi 214122, China\\
         3. Kyushu Institute of Technology, Fukuoka 8048550, Japan\\
$^*$Corresponding author. E-mail Address: zhujh@xjtu.edu.cn}

\begin{abstract}
This paper proposes an effective approach for the scaling registration of $m$-D point sets. Different from the rigid transformation, the scaling registration can not be formulated into the common least square function due to the ill-posed problem caused by the scale factor. Therefore, this paper designs a novel objective function for the scaling registration problem. The appearance of this objective function is a rational fraction, where the numerator item is the least square error and the denominator item is the square of the scale factor. By imposing the emphasis on scale factor, the ill-posed problem can be avoided in the scaling registration. Subsequently, the new objective function can be solved by the proposed scaling iterative closest point (ICP) algorithm, which can obtain the optimal scaling transformation. For the practical applications, the scaling ICP algorithm is further extended to align partially overlapping point sets. Finally, the proposed approach is tested on public data sets and applied to merging grid maps of different resolutions. Experimental results demonstrate its superiority over previous approaches on efficiency and robustness.
\end{abstract}

\end{frontmatter}



\section{Introduction}

Due to the wide application in 3D reconstruction \citep{Arrig16,Zhu16}, shape recognition \citep{Kakad07,Abate08}, and robot mapping \citep{Ma16}, point sets registration has attracted immense attention \cite{Tam13} in computer vision, pattern recognition, and robotics. The goal of registration is to establish correspondences between two point sets and recover the optimal transformation, which can well align one point set to the other. It has been the object of much attention since the seminal work presented in \cite{Besl92,Chen92}.

To solve the registration problem, the most popular solution is the iterative closest point (ICP) algorithm \cite{Besl92}. In this approach, the rigid registration problem can be formulated into a least square (LS) function, which can be solved by establishing the point correspondences and calculating the rigid transformation iteratively. Although the ICP algorithm can achieve the rigid registration with good accuracy and efficiency, it cannot be applied to the registration of partially overlapping point sets. To trim the outliers, Chetverikov et al. \cite{Chet05} introduced an overlap parameter into the LS function, which can be solved by the trimmed ICP (TrICP) algorithm. By discarding outliers, this approach can obtain accurate registration results for partially overlapping point sets. Nevertheless, it is time-consuming to calculate the overlap parameter. To accelerate the registration, Phillips et al. \cite{Phil07} proposed the fractional TrICP (FTrICP) algorithm, which can compute the overlap parameter and the optimal transformation simultaneously. As the variants of ICP algorithm are well known to be susceptible to local convergence, the particle filter \cite{Sandhu10} or genetic algorithm \cite{Lomo06} could be utilized to obtain the desired global minimum. Besides, effective features \cite{Lowe04,Rusu09,Lei17} can also be extracted and matched for the point sets to be registered so as to provide the initial parameters for registration approaches. These registration approaches may obtain accurate results for the rigid registration, but they may not achieve the scaling registration, which may exist in some practical applications.

To achieve the fine registration, Zha et al. \cite{Zha00} integrated the scale factor into a modified ICP algorithm, where extended signatures images were utilized to establish precise correspondence. Besides, Zinber et al. \cite{Zin05} also proposed to integrate the scale factor into the LS function, where the scale factor is directly estimated. But it can always obtain the unexpected registration results due to ill-posed problem inherited in the scaling registration. Therefore, Ying et al. \cite{Ying09} proposed an approach by applying the boundary constraint as a regularizer to the scale factor in the LS function and it may obtain accurate registration results. However, it is difficult to estimate the suitable range of scale factor in some cases, which may lead to the failure of registration. Subsequently, Zhu et al. \cite{Zhu10} introduced the bidirectional distance measurement into the scaling registration and designed the corresponding LS function, which can be solved by the proposed scaling ICP algorithm. What's more, Du et al. \cite{Du11} extended the bidirectional distance measurement to deal with the scaling registration of partially overlapping point sets. Although this approach is very robust, it is somewhat time-consuming due to the establishment of bidirectional correspondences.

Meanwhile, some probabilistic approaches were also proposed for the scaling registration of partially overlapping point sets. By representing range images by Gaussian mixture models (GMMs), Myronenko and Song \citep{Song2010} considered the alignment of two range images as a probability density estimation problem, which can be solved by iteratively fitting the
GMM centroids with EM algorithm. In \citep{Bing11}, range images are also represented as GMMs, then the registration
problem is viewed as that of aligning two Gaussian mixtures by minimizing their discrepancy. Although these probabilistic approaches are very accurate, they are always requiring huge computational resources. Recently, Du et al. \cite{Du16} combined the ICP algorithm with EM algorithm to achieve the scaling registration. It also needs to estimate boundaries of the scale factor. Therefore, a more effective approach is required to be proposed for the scaling registration of point sets.

Compared to the rigid registration, the scale factor is the extra item in the scaling registration, which can increase the difficulty of this problem. As the criteria of least square (LS) error is an ill-posed problem in the scaling registration, we should pay more attention to the scale factor. By imposing the emphasis on the scale factor, we can design a new objective function, where the ill-posed problem can be transformed into the well-posed one for the scaling registration of absolutely overlapping point sets. Subsequently, the solution of this objective function will be proposed to obtain the optimal scaling transformation. For the practical applications, the proposed approach is then extended to deal with partially overlapping point sets. Besides, it will be applied to merging grid maps of different resolutions.

The remainder of this paper is organized as follows. Section 2 briefly reviews the original ICP algorithm. In Section 3, the proposed approach is presented for the scaling registration of $m$-D point sets. Following that is Section 4, in which the proposed approach is tested and evaluated on some public data sets. Finally, some conclusions are drawn in Section 5.

\section{The ICP algorithm}

Given two overlapping point sets in ${\mathbb{R}^m}$, the data shape $P =\{ {\vec p_i}\} _{i = 1}^{{N_p}}$ and the model shape $Q =\{ {\vec q_j}\} _{j = 1}^{{N_q}}({N_p},{N_q}\in{\mathbb{N}})$, the goal of rigid registration is to find the optimal transformation $({\bf{R}},\vec t)$, with which $P$ can be in the best alignment with $Q$. Accordingly, it can be formulated as the following LS problem:	
\begin{equation}
\begin{array}{*{20}{c}}
   {\mathop {\min }\limits_{\;{\bf{R}},\vec t,c(i) \in \{ 1,...,{N_q}\} } \sum\limits_{i = 1}^{{N_p}} {\left\| {{\bf{R}}{{\vec p}_i} + \vec t - {{\vec q}_{c(i)}}} \right\|_2^2} }  \\
   {{\rm{s}}{\rm{.t}}{\rm{.}}{{\bf{R}}^T}{\bf{R}} = {{\bf{I}}_{m}},\det ({\bf{R}}) = 1}  \\
\end{array},
\label{eq:LS}
\end{equation}
where ${\bf{R}} \in \mathbb{R}^{m\times m}$ is the rotation matrix, $\vec t \in {\mathbb{R}^m}$ indicates the translation vector, ${\vec q_{c(i)}}$ represents the correspondence of the $i$th point ${\vec p_i}$ in the model shape, and ${\left\| \right\|}_2$ denotes the $L_2$ norm. Eq. (\ref{eq:LS}) can be solved by the ICP algorithm \cite{Besl92}, which can achieve the rigid registration by iterations. Given the initial transformation $({\bf{R}_0},{\vec t_0})$, two steps are included in each iteration:

(1) Establish new correspondence for each point ${{\vec p}_i}$ in the data shape $P$:
\begin{equation}
{c_k}(i) = \mathop {\arg \min }\limits_{j \in \{ 1,2,...,{N_q}\} } {\left\| {{\bf{R}_{k - 1}}{{\vec p}_i} + {{\vec t}_{k - 1}} - {{\vec q}_j}} \right\|_2}
\end{equation}

(2) Update the rigid transformation by minimizing the following function:
 \begin{equation}
 ({{\bf{R}}_k},{\vec {t}_k}) = \mathop {\arg \min }\limits_{{\bf{R}},\vec t} \sum\limits_{i = 1}^{{N_p}} {\left\| {{\bf{R}}{{\vec p}_i} + \vec t - {{\vec q}_{{c_k}(i)}}} \right\|_2^2}
 \end{equation}

Although the original ICP algorithm has good performance, it cannot deal with the scaling registration problem. Besides, it
does not fit for the registration of partially overlapping point sets.

\section{The proposed approach}
Suppose there are two overlapping point sets in ${\mathbb{R}^m}$, the data shape $P =\{ {\vec p_i}\} _{i = 1}^{{N_p}}$ and the model shape $Q =\{ {\vec q_j}\} _{j = 1}^{{N_q}}({N_p},{N_q}\in{\mathbb{N}})$.
Accordingly, the goal of scaling registration to find the optimal scaling transformation $(s,{\bf{R}},\vec t)$, with which $P$ can be in the best alignment with $Q$. Therefore, it seems that this registration problem can reasonably be formulated as the following LS function:	
\begin{equation}
\begin{array}{l}
 \mathop {\min }\limits_{s{\rm{,R}},\vec t,j \in \{ 1,...,{N_q}\} } (\sum\limits_{i = 1}^{{N_p}} {\left\| {s{\bf{R}}{{\vec p}_i} + \vec t - {{\vec q}_{c(i)}}} \right\|_2^2} ). \\
 {\rm{s}}{\rm{.t}}{\rm{.}}\quad \quad {{\bf{R}}^{\rm{T}}}{\bf{R}}{\rm{ = }}{{\rm{I}}_d},\det ({\bf{R}}) = 1,s > 0 \\
 \end{array}
 \label{eq:LSs}
 \end{equation}
where $s$ denotes the scale factor.
\begin{figure}[htp]
\begin{center}
\subfigure[]{\label{fig:demo-a}\includegraphics[scale=0.9]{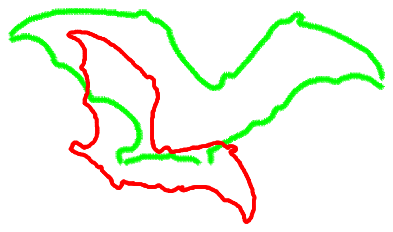}}
\subfigure[]{\label{fig:demo-b}\includegraphics[scale=0.9]{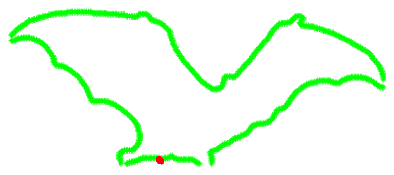}}
\subfigure[]{\label{fig:demo-c}\includegraphics[scale=0.9]{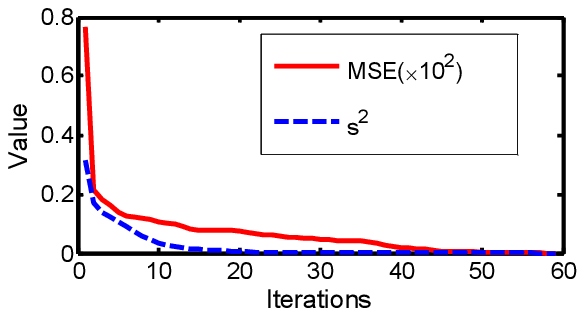}}
\end{center}
\caption{The illustration of phenomenon with $s \to 0$. (a) The data shape (red) and model shape (green) to be aligned. (b) The undesirable result of scaling registration. (c) The value of two items change in each iteration.}
\label{fig:demo}
\end{figure}

Although Eq. (\ref{eq:LSs}) is the formulation of the scaling registration, it is unreasonably adopted as the objective function of this registration problem due to one reason. As shown in Fig.\ref{fig:demo}, if the scale factor $s$ tends to zero, there will appear the phenomenon that all points in data shape $P$ will be close to each other and find the same corresponding point in model shape $Q$. And when $s=0$, the objective function denoted by Eq. (\ref{eq:LSs}) can be minimized. But this is not the expected registration result. Therefore, Eq. (\ref{eq:LSs}) is an ill-posed problem, which can lead to unexpected registration results.

To address this issue, a novel objective function can be proposed by imposing the emphasis on the scale factor as follows:
\begin{equation}
\begin{array}{l}
 \mathop {\min }\limits_{s{\rm{,R}},\vec t,j \in \{ 1,...,{N_q}\} } (\sum\limits_{i = 1}^{{N_p}} {\left\| {s{\bf{R}}{{\vec p}_i} + \vec t - {{\vec q}_{c{j}}}} \right\|_2^2/{s^v}} ). \\
 {\rm{s}}{\rm{.t}}{\rm{.}}\quad \quad {{\bf{R}}^{\rm{T}}}{\bf{R}}{\rm{ = }}{{\rm{I}}_m},\det ({\bf{R}}) = 1 \\
 \end{array}
 \label{eq:LSn}
\end{equation}
As Eq. (\ref{eq:LSn}) and Fig. \ref{fig:demo-c} displayed, if the scale factor $s$ tends to zero, the item of denominator will be decreased to zero quickly, even when the item of the numerator may be gradually declined, which can avoid the ill-posed problem appeared in Eq. (\ref{eq:LSs}).

Here, the value of $v$ in the denominator item should be carefully chosen due to two reasons. Firstly, the item denoted by the denominator should be decreased more rapidly than or equal to that of the numerator, when the scale factor tends to zeros. Otherwise, it could not avoid the phenomenon that the scale factor tends to zero. As the item of the numerator denotes the quadratic error, the value of $v$ must be greater than or equal to 2. Secondly, it is required that Eq. (\ref{eq:LSn}) should contain the valid solution, which can be easily obtained. During the derivation, we find $v$=1 or $v$=2 can make it easy to obtain the solution, which is valid. Therefore, we can only choose $v$=2 in Eq. (\ref{eq:LSn}).

\subsection{The scaling ICP algorithm}

According to the original ICP algorithm, Eq. (\ref{eq:LSn}) can be solved by the proposed scaling ICP algorithm, which
achieves the scaling registration by iterations. Given the initial transformation $(s_0, {\bf{R}_0},{\vec t_0})$, two steps are included in each iteration:

(1) Establish correspondences between two $m$-D point sets:
\begin{equation}
{c_k}(i) = \mathop {\arg \min }\limits_{j \in \{ 1,2,..,{N_q}\} } {\left\| {{s_{k - 1}}{{\bf{R}}_{k - 1}}{{\vec p}_i} + {{\vec t}_{k - 1}} - {{\vec q}_j}} \right\|_2}\quad {\rm{for}}\;i = 1,2,...,{N_p}
\end{equation}

(2) Calculate the scaling transformation by minimizing the function:
\begin{equation}
({s_k},{{\bf{R}}_k},{\vec t_k})\mathop { = \arg \min }\limits_{s,{\bf{R}},\vec t} \frac{{\sum\nolimits_{i = 1}^{_{_{{N_p}}}} {\left\| {s{\bf{R}}{{\vec p}_i} + \vec t - {{\vec q}_{{c_k}(i)}}} \right\|_2^2} }}{{{s^2}}}
\end{equation}

Step (1) can be solved by many efficient methods such as the nearest neighbor search method based on the $k$-d tree or its variants \cite{Nuch07}. Therefore, Step (2) is the critical step.

To calculate the scaling transformation, the following lemma can be presented without proof.

{\bfseries{Lemma 1}} Given two $m$-D point sets, $D \buildrel \Delta \over = \{ {\vec d_i}\} _{i = 1}^N$ and $M \buildrel \Delta \over = \{ {\vec m_i}\} _{i = 1}^N$, then the function $F(t) = \sum\nolimits_{i = 1}^N {\left\| {{{\vec d}_i} + \vec t - {{\vec m}_i}} \right\|_2^2}$ has the minimum value when $\vec t = \frac{1}{N}\sum\nolimits_{i = 1}^N {{{\vec m}_i}}  - \frac{1}{N}\sum\nolimits_{i = 1}^N {{{\vec d}_i}}$.

According to this lemma, minimizing $F(s,{\bf{R}}) = \sum\nolimits_{i = 1}^{_{_{{N_p}}}} {\left\| {s{\bf{R}}{{\vec p}_i} + \vec t - {{\vec q}_{{c_k}(i)}}} \right\|_2^2/{s^2}}$ can obtain the following result:
\begin{equation}
\vec t = \frac{1}{N}\sum\limits_{i = 1}^{{N_p}} {{{\vec q}_{{c_k}(i)}} - \frac{1}{N}} \sum\limits_{i = 1}^{{N_p}} {s{\bf{R}}{{\vec p}_i}}.
\label{eq:t}
\end{equation}
Hence, the function $F(s,{\bf{R}})$ can be transformed into the following form:
\begin{equation}
F(s,{\bf{R}}) = \sum\limits_{i = 1}^{_{_{{N_p}}}} {\left\| {s{\bf{R}}({{\vec p}_i} - \frac{1}{N}\sum\nolimits_{i = 1}^{{N_p}} {{{\vec p}_i}} ) - ({{\vec q}_{{c_k}(i)}} - \frac{1}{N}\sum\nolimits_{i = 1}^{{N_p}} {{{\vec q}_{{c_k}(i)}}} )} \right\|_2^2/{s^2}}.
\end{equation}

Denote ${\vec d_i} = {\vec p_i} - \frac{1}{{{N_p}}}\sum\nolimits_{i = 1}^{{N_p}} {{{\vec p}_i}}$ and ${\vec m_i} = {\vec q_{{c_k}(i)}} - \frac{1}{{{N_p}}}\sum\nolimits_{i = 1}^{{N_p}} {{{\vec q}_{{c_k}(i)}}}$. Then, $ F(s,{\bf{R}})$ can be simplified and expanded as:
\begin{equation}
\begin{array}{l}
 F(s,{\bf{R}}) = \sum\limits_{i = 1}^{{N_p}} {\left\| {s{\bf{R}}{{\vec d}_i} - {{\vec m}_i}} \right\|_2^2} /{s^2} \\
 \quad \quad \quad  = ({s^2}\sum\limits_{i = 1}^{{N_p}} {\vec d_i^T} {{\vec d}_i} - 2s\sum\limits_{i = 1}^{{N_p}} {\vec d_i^T} {\bf{R}}{{\vec m}_i} + \sum\limits_{i = 1}^{{N_p}} {\vec m_i^T} {{\vec m}_i})/{s^2}. \\
 \end{array}
 \label{eq:stran}
 \end{equation}
Subsequently, the scaling transformation can be calculated as follows.

(a) Rotation matrix

As shown in Eq. (\ref{eq:stran}), only the middle item of $F(s,{\bf{R}})$ contains the rotation matrix ${\bf{R}}$.
Therefore, the function $F(s,{\bf{R}})$ can be minimized by maximizing the item $s\sum\nolimits_{i = 1}^{{N_p}} {\vec d_i^T} {\bf{R}}{\vec m_i}$. As $s$ is a scale factor, maximizing the item $s\sum\nolimits_{i = 1}^{{N_p}} {\vec d_i^T} {\bf{R}}{\vec m_i}$ is equivalent to maximizing the item $\sum\nolimits_{i = 1}^{{N_p}} {\vec d_i^T} {\bf{R}}{\vec m_i}$ for calculating the rotation matrix. According to \cite{Nuch10}, the matrix H with its SVD can be computed as follows:
\begin{equation}
    {\bf{H}} = \frac{1}{{{N_p}}}\sum\limits_{i = 1}^{{N_p}} {\vec d_i^T} {\vec m_i}
\end{equation}
\begin{equation}
U\Lambda V= svd({\bf{H}})
\end{equation}
where $svd(.)$ denotes the function of singular value decomposition. Then, the rotation matrix can
be obtained as follows:
\begin{equation}
{{\bf{R}}_k} = V{U^T}
\end{equation}

(b) Scale factor

Taking the derivative of $F(s,{\bf{R}})$ with respective to the scale factor $s$, it is easy to get
the following result:
\begin{equation}
\frac{{\partial F(s,{\bf{R}})}}{{\partial s}} = 2{s^{ - 2}}\sum\limits_{i = 1}^{{N_p}} {\vec d_i^T} {\bf{R}}{\vec m_i} - 2{s^{ - 3}}\sum\limits_{i = 1}^{{N_p}} {\vec m_i^T} {\vec m_i}.
\end{equation}
Let $\frac{{\partial F(s,{\bf{R}})}}{{\partial s}} = 0$, the scale factor can be calculated as follows:
\begin{equation}
s_k = \frac{{\sum\nolimits_{i = 1}^{{N_p}} {\vec m_i^T{{\vec m}_i}} }}{{\sum\nolimits_{i = 1}^{{N_p}} {\vec d_i^T{{\bf{R}}_k}{{\vec m}_i}} }}
\end{equation}

Obviously, if the value of $n$ is set to be greater than 2 in Eq. (\ref{eq:LSn}), there are three items in ${\partial F(s,{\bf{R}})/\partial s}$. Accordingly, the scale factor $s$ will exit more than one solution, which may be the complex number. This is invalid and unexpected.

(c) Translation vector

After getting the rotation matrix and scale factor, the translation vector can be calculated by Eq. (\ref{eq:t}).
\begin{equation}
\vec t = \frac{1}{N}\sum\limits_{i = 1}^{{N_p}} {{{\vec q}_{{c_k}(i)}} - \frac{1}{N}} \sum\limits_{i = 1}^{{N_p}} {s_k{{\bf{R}}_k}{{\vec p}_i}}
\end{equation}

By repeating Step(1)$\sim$(2), the scaling ICP algorithm can achieve the scaling registration of absolutely overlapping point sets.

\subsection{The scaling and trimmed ICP algorithm}

In many practical applications, point sets to be registered are always partially overlapping. Therefore, the proposed scaling registration algorithm should be extended to align partially overlapping point sets.

Suppose there are two overlapping overlapping point sets in $\mathbb{R}^m$, the model shape $Q \buildrel \Delta \over = \{ {\vec q_i}\} _{i = 1}^{{N_q}}$ and a data shape $P \buildrel \Delta \over = \{ {\vec p_i}\} _{i = 1}^{{N_p}}$ $({N_p},{N_q} \in \mathbb{N})$, where $\xi$ represents the overlapping percentage of the data shape, ${P_\xi }$ denotes the point subset, which is the overlapping part of the data shape to the model shape and $N_p^{'}$ indicates the number of points in ${P_\xi }$. According to the idea of the trimmed ICP algorithm \cite{Chet05}, the optimal scaling transformation can be obtained by minimizing the following objective function:
\begin{equation}
\Psi (\xi ,s,{\bf{R}},\vec t) = \frac{{e(\xi ,s,{\bf{R}},\vec t)}}{{{s^2} \cdot {\xi ^{1 + \lambda }}}},
\label{eq:sTrICP}
\end{equation}
where
\begin{equation}
e (\xi ,s,{\bf{R}},\vec t) = \frac{{\sum\limits_{{{\vec p}_i} \in {P_\xi }} {\left\| {s{\bf{R}}{{\vec p}_i} + \vec t - {{\vec q}_{c(i)}}} \right\|_2^2} }}{{ N_p^{'}}}.
\end{equation}

To solve Eq. (\ref{eq:sTrICP}), the scaling and trimmed ICP (sTrICP) algorithm can be proposed to achieve the scaling registration by iterations. Given the initial transformation $(s_0, {\bf{R}_0},{\vec t_0})$, three steps are included in each iteration:

(1) Establish the point correspondence:
\begin{equation}
{c_k}(i) = \mathop {\arg \min }\limits_{j \in \{ 1,2,..,{N_q}\} } {\left\| {{s_{k - 1}}{{\bf{R}}_{k - 1}}{{\vec p}_i} + {{\vec t}_{k - 1}} - {{\vec q}_j}} \right\|_2}\quad {\rm{for}}\;i = 1,2,...,{N_p}.
\label{eq:sTrICP1}
\end{equation}

(2) Update the overlapping percentage and its corresponding subset:
\begin{equation}
({\xi _k},{P_{{\xi _k}}})\mathop { = \arg \min }\limits_{\xi ,{P_{{\xi}}}} \Psi (\xi ,{s},{{\bf{R}}},{\vec t}).
\label{eq:sTrICP2}
\end{equation}

(3) Calculate the scaling transformation:
\begin{equation}
({s_k},{{\bf{R}}_k},{\vec t_k})\mathop { = \arg \min }\limits_{s,{\bf{R}},\vec t} \sum\limits_{{{\vec p}_i} \in {P_{\xi_k} }} {\left\| {s{\bf{R}}{{\vec p}_i} + \vec t - {{\vec q}_{c_k(i)}}} \right\|_2^2/{s^2}}.
\label{eq:sTrICP3}
\end{equation}

Obviously, Step (1) and (3) are similar to that of the proposed scaling registration algorithm. And Step (2) can be solved in a sequence processing manner. Denote ${\vec p_{i,k}} = {s_{k - 1}}{{\bf{R}}_{k - 1}}{\vec p_i} + {\vec t_{k - 1}}$, then the point pairs $\{ {\vec p_{i,k}},{\vec q_{{c_k}(i)}}\} _{i = 1}^{{N_p}}$  can be sorted by their distances in the ascending order. Each time, a pair of sorted points can be added to compute the corresponding value of $\Psi (\xi )$. By traveling all sorted point pairs, it is easy to obtain the minimum value $\Psi ({\xi _k})$, which corresponds to the optimal overlapping percentage ${\xi _k}$. Then, the points involved in the front ${\xi _k}{N_p}$ sorted point pairs can be selected to update the corresponding point subset ${P_{{\xi _k}}}$.

By repeating Step(1)$\sim$(3), the scaling and trimmed ICP algorithm can achieve the scaling registration of partially overlapping point sets.

\subsection{Proof of Convergence}

As the sTrICP algorithm is proposed based on the idea of the original ICP algorithm, they have the similar convergence property. The following theorem will explain this convergence in detail so as to prove that this algorithm converges in theory.

{\bfseries{Theorem.}} The sTrICP algorithm converges monotonically to a local minimum with respect to the objective function value.

The following proof shows that, for each step of the TsICP algorithm, the value of the objective function can be no worse than the previous step.

{\bfseries{Proof.}} Suppose there are two partially overlapping point sets, $P \buildrel \Delta \over = \{ {\vec p_i}\} _{i = 1}^{{N_p}}$ and $Q \buildrel \Delta \over = \{ {\vec q_i}\} _{i = 1}^{{N_q}}$. Denote ${T_k} = ({s_k},{{\bf{R}}_k},{\vec t_k})$, ${\xi _k}$ and ${P_{{\xi _k}}}$ as the scaling transformation, overlapping percentage, and overlapping part of $P$ to $Q$, respectively. In the first step of the $k$th iteration, the nearest neighbour ${\vec q_{{c_k}(i)}}$ in $Q$ is searched for the point ${\vec p_i}$ in $P$. Let ${\vec p_{i,k - 1}} = {s_{k - 1}}{{\bf{R}}_{k - 1}}{\vec p_i} + {\vec t_{k - 1}}$. The value of objective function can then be defined as:
\begin{equation}
{e_k} = \frac{{\sum\limits_{{{\vec p}_i} \in {P_{{\xi _{k - 1}}}}} {\left\| {{{\vec p}_{i,k - 1}} - {{\vec q}_{{c_k}(i)}}} \right\|_2^2} }}{{\left| {{P_{{\xi _{k - 1}}}}} \right| \cdot {s_{k - 1}}^2 \cdot {{({\xi _{k - 1}})}^{1 + \lambda }}}}.
\end{equation}
In the second step of the $k$th iteration, the percentage ${\xi _k}$ and corresponding subset ${P_{{\xi _k}}}$ are updated.
Then, the updated value of objective function is:
\begin{equation}
{\eta _k} = \frac{{\sum\limits_{{{\vec p}_i} \in {P_{{\xi _k}}}} {\left\| {{{\vec p}_{i,k - 1}} - {{\vec q}_{{c_k}(i)}}} \right\|_2^2} }}{{\left| {{P_{{\xi _k}}}} \right| \cdot {s_{k-1}}^2 \cdot {{({\xi _k})}^{1 + \lambda }}}}
\end{equation}
As ${\xi _k}$ and ${P_{{\xi _k}}}$ are calculated from Eq. (\ref{eq:sTrICP2}), it is reasonable to conclude that ${\eta _k} \le {e_k}$. In the third step of the $k$th iteration, $\{ {\vec q_{{c_k}(i)}}\} _{i = 1}^{\left| {{P_{{\xi _k}}}} \right|}$ is registered with $\{ {\vec p_{i,k - 1}}\} _{i = 1}^{\left| {{P_{{\xi _k}}}} \right|}$ and the scaling transformation ${{T}_k}{\rm{ = (}}{s_k}{\rm{,}}{{\bf{R}}_k},{\vec t_k})$ is optimized. Let ${\vec p_{i,k}} = {s_k}{{\bf{R}}_k}{\vec p_i} + {\vec t_k}$. And the value of objective function becomes:
\begin{equation}
{\varepsilon _k} = \frac{{\sum\limits_{{{\vec p}_i} \in {P_{{\xi _k}}}} {\left\| {{{\vec p}_{i,k}} - {{\vec q}_{{c_k}(i)}}} \right\|_2^2} }}{{\left| {{P_{{\xi _k}}}} \right| \cdot {s_k}^2 \cdot {{({\xi _k})}^{1 + \lambda }}}}.
\end{equation}
Since the scaling transformation ${{T}_k}{\rm{ = (}}{s_k}{\rm{,}}{{\bf{R}}_k},{\vec t_k})$ is computed from Eq. (\ref{eq:sTrICP3}), it is easy to conclude that ${\varepsilon _k} \le {\eta _k}$.

In the $(k+1)$th iteration, the new nearest neighbour ${\vec q_{{c_{k + 1}}(i)}}$ can be searched for ${\vec p_i}$ again. Subsequently, the new value of objective function can be defined as:
\begin{equation}
{e_{k + 1}} = \frac{{\sum\limits_{{{\vec p}_i} \in {P_{{\xi _k}}}} {\left\| {{{\vec p}_{i,k}} - {{\vec q}_{{c_{k + 1}}(i)}}} \right\|_2^2} }}{{\left| {{P_{{\xi _k}}}} \right| \cdot {s_k}^2 \cdot {{({\xi _k})}^{1 + \lambda }}}}.
\end{equation}
Because $e_{k + 1}$ is the result updated from the Eq. (\ref{eq:sTrICP1}), therefore:
\begin{equation}
{e_{k + 1}} = \frac{{\sum\limits_{{{\vec p}_i} \in {P_{{\xi _k}}}} {\left\| {{{\vec p}_{i,k}} - {{\vec q}_{{c_{k + 1}}(i)}}} \right\|_2^2} }}{{\left| {{P_{{\xi _k}}}} \right| \cdot {s_k}^2 \cdot {{({\xi _k})}^{1 + \lambda }}}} \le \frac{{\sum\limits_{{{\vec p}_i} \in {P_{{\xi _k}}}} {\left\| {{{\vec p}_{i,k}} - {{\vec q}_{{c_k}(i)}}} \right\|_2^2} }}{{\left| {{P_{{\xi _k}}}} \right| \cdot {s_k}^2 \cdot {{({\xi _k})}^{1 + \lambda }}}} = {\varepsilon _k}
\end{equation}
Hence, repeating the above procedures, the following results can be reasonably obtained:

$0 \le ... \le {e_{k + 1}} \le {\varepsilon _k} \le {\eta _k} \le {e_k} \le ...$ \quad for all $k$

According to the Monotonic Sequence Theorem: every bounded monotonic sequence of real numbers is convergent, we can reasonably conclude that the sTrICP algorithm always converges monotonically to a local minimum with respect to the objective function value.

\section{Experimental results}

In this section, experiments were performed on two public available datasets:
(1) 2D shapes in part B of CE-Shape \cite{Late00}, (2) the Stanford 3D Scanning Repository \cite{Stanford}.

To demonstrate its performance, the proposed approaches (sTrICP)
was compared with the ICP algorithm with bounded scale \cite{Ying09} and the ICP algorithm
based on the bidirectional distance measurement \cite{Du11}, which are abbreviated to TrICPBs and BiTrICP, respectively.
The reason for the comparison with these two algorithms is that all these approaches are proposed under the framework of the original ICP algorithm. As the performance of TrICPBs is relatively affected by the boundaries of scale factor, the Principal Component Analysis (PCA) method was adopted to estimate its initial value $s_0$. For the comprehensive comparison,
two groups of boundaries: $[0.9{s_0},1.1{s_0}]$ and $[0.5{s_0},2{s_0}]$ are both applied to the TrICPBs algorithm.
For distinction, the TrICPBs algorithm with narrow and wide boundaries are abbreviated to TrICPBNs and TrICPBWs, respectively.

All the competing approaches utilized the nearest neighbor search method based on $k$-d tree to establish correspondences.
Besides, no change of all point correspondences and the maximum iterations are both adopted as the termination condition of iterations for these completed approaches. Experiments were performed in MATLAB and conducted on a laptop with 2.5GHz processor of double-cores and 4GB RAM.

\subsection{2D shapes}

\begin{figure}
\begin{center}
\includegraphics[width= 0.5\linewidth]{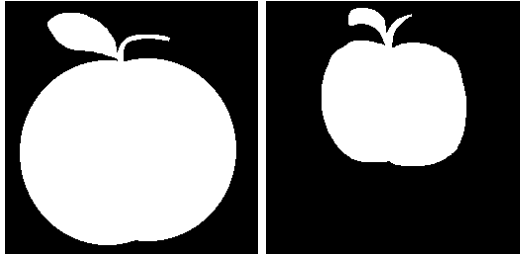}
\end{center}
   \caption{The original shapes selected from Part B of CE-Shape-1.}
\label{fig:app}
\end{figure}

Here, experiments were conducted to illustrate the performance of the proposed approach for the scaling registration of 2D shapes. The data sets were selected from Part B of CE-Shape-1, which is a large 2D shapes database. Before the experiment, one shape was selected as the data shape and the other one was viewed as the model shape. Fig. \ref{fig:app} displays Apple shapes selected for the experiment. To generate partially overlapping point sets, some selected shapes should be cut by one part. Then, edge points were extracted for each shapes so as to obtain two point sets to be registered. Besides, a random scale factor was generated and might be applied to the data shape. Accordingly, all competing approaches can be applied to the registration of these shape pairs.

During experiment, the runtime required for the registration of each range image pair was recorded for all competing approaches.
As these registration approaches are all efficient, 20 Monte Carlo (MC) trials were carried
out for each shape pair so as to eliminate randomness. For comparison, Table \ref{Tab:2D} records the
objective function value and average runtime for all these competing approaches.

\begin{table}
\caption{Performance comparison of competing approaches tested on different 2D shapes}
 \centering \scriptsize
\centering
\begin{tabular}{|c|c|c|c|c|c|c|c|c|c|}
\hline
&\multirow{2}{*}{} & \multicolumn{2}{|c|}{BiTrICP} & \multicolumn{2}{|c|}{TrICPBNs} & \multicolumn{2}{|c|}{TrICPBWs} & \multicolumn{2}{|c|}{sTrICP}\\
\cline{3-10}
Datasets &$s$   & MSE    & T(s)  & MSE    & T(s)    & MSE      & T(s)   & MSE   & T(s) \\
\hline
Apple    &1.4733      &1.4508  &0.1893 &0.8277  &0.0436	&0.8277	   &0.0466	&0.8275	&0.0352\\
\hline
 Bat     &0.75	    &0.2682  &0.8372 &7.0585  &0.1979	&196.0914  &0.3584	&0.2680	&0.3771\\
\hline
Chicken  &1.5	    &0.2668  &0.3368 &0.2675  &0.1741	&1.7866	   &0.1896	&0.2675	&0.0755\\
\hline
\end{tabular}
\label{Tab:2D}
\end{table}

\begin{figure}
\begin{center}
\includegraphics[width= 0.6\linewidth]{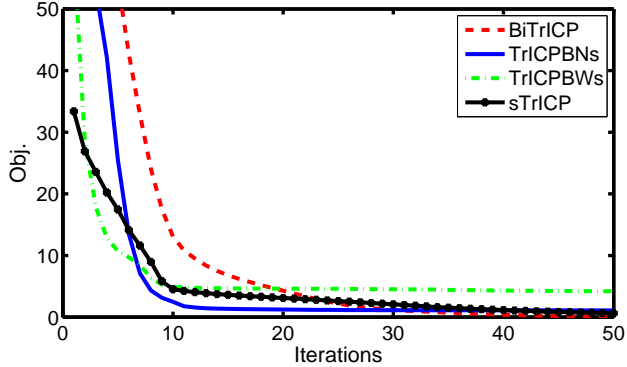}
\end{center}
   \caption{The convergence of all competing approaches tested on Apple shape.}
\label{fig:2DCon}
\end{figure}

As shown in Fig. \ref{Tab:2D}, all competing approaches can obtain accurate results for the Apple shapes. Therefore, Fig. \ref{fig:2DCon} illustrates the change of objective function value for all competing approaches tested on  this shape. As depicted in Fig. \ref{fig:2DCon}, the sTrICP algorithm is locally convergent for the registration of 2D point sets. Besides, all these registration approaches converge in the similar way with a fast speed. This is because all these registration approaches are proposed under the framework of the original ICP algorithm, which can achieve registration with fast speed.

For other shapes, both the BiTrICP and sTrICP algorithms can also achieve the scaling registration with good accuracy.
By introducing the bidirectional measurement, the BiTrICP algorithm can avoid the ill-posed problem involved in the scaling registration. Therefore, this approach is very robust. But it is somewhat time-consuming due to the establishment of the bidirectional correspondences. While, the sTrICP algorithm is proposed the new objective function, which imposes the emphasis on the scale factor. As shown in the Fig. \ref{fig:demo-c}, when the scale factor converges to 0, the mean square error gradually turns to be small and the value of the item $s^2$ reduces rapidly. Therefore, the objective function become large. Since the scaling transformation can be calculated by minimizing the well-designed objective function, the emphasis on the scale factor can avoid the ill-posed problem involved in the scaling registration. Obviously, there is no need to limit the boundaries of the scale factor. What's more, it only required to establish the forward correspondence. Therefore, the proposed approach is also very robust and more efficient than that of the BiTrICP algorithm.

However, the TrICPBs algorithm may not always obtain good results for the scaling registration of 2D shape pairs. To avoid the ill-posed problem, the TrICPBs algorithm suggests that the boundaries of the scale factor should be limited due to its initial value. As displayed in the second row of Fig. \ref{fig:2DRes}, for partially overlapping point sets, it is difficult to estimate the initial value of the scale factor. To cover the real scale factor, the scale factor should be limited with a wide range. In this case, the scale factor also will converge to the direction of 0 and trapped into a local minimum around the low boundary. But if the scale factor is limited with a narrow range, which may not cover its real value, the TrICPBs approach will never obtain the accurate results for the scaling registration.

Therefore, the proposed approach can achieve the scaling registration of 2D shapes with superior performances than other two related approaches.

\subsection{Range images}

Here, experiments were conducted to illustrate the performance of the proposed approach for scaling registration of range images. The data sets were selected from the Stanford 3D Scanning Repository. Before experiment, one range image was selected as the data shape and the other one was viewed as the model shape. Besides, a random scale factor was generated and applied to the data shape. Subsequently, all competing approaches can be applied to registration of these range image pairs.

During experiment, the runtime required for the registration of each range image pair was recorded for all competing approaches.
As all competing approaches are efficient, 10 MC trials were carried
out for each range image pair so as to eliminate randomness. For comparison, Table \ref{Tab:3D} records the
objective function value and average runtime for all these competing approaches.

\begin{table}
\caption{Performance comparison of all competing approaches tested on different range images}
\centering \scriptsize
\centering
\begin{tabular}{|c|c|c|c|c|c|c|c|c|c|}
\hline
&\multirow{2}{*}{} & \multicolumn{2}{|c|}{BiTrICP} & \multicolumn{2}{|c|}{TrICPBNs} & \multicolumn{2}{|c|}{TrICPBWs} & \multicolumn{2}{|c|}{sTrICP}\\
\cline{3-10}
Datasets    &$s$     & MSE   & T(s)  & MSE   & T(s)    & MSE     & T(s)  & MSE   & T(s) \\
\hline
Bunny       & 0.5	&1.2676	&3.7437	&1.2747	&1.1191	&4.1380	  &2.9040	&1.2698	&1.6219\\
\hline
Dragon      &0.25	&0.5560	&3.4324	&9.0729	&4.3301	&18.2296  &1.8909	&0.5539	&2.2583\\
\hline
HappyBuddha &1.25	&0.6266	&5.3875	&0.6206	&2.0002	&0.6206	  &1.9986	&0.6280	&1.9558	\\
\hline
\end{tabular}
\label{Tab:3D}
\end{table}

\begin{figure}
\begin{center}
\includegraphics[width= 0.6\linewidth]{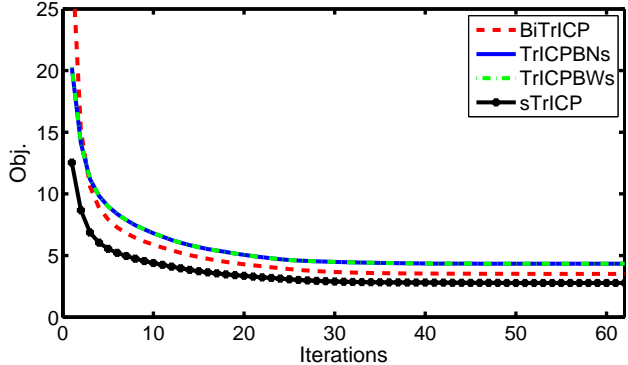}
\end{center}
   \caption{The convergence of all competing approaches tested on the Stanford Happy Buddha.}
\label{fig:3DCon}
\end{figure}

As shown in Fig. \ref{Tab:3D}, all competing approaches can obtain accurate results for the Stanford Happy Buddha. Therefore, Fig. \ref{fig:2DCon} illustrates the change of objective function value for all all competing approaches tested on this data set. As depicted in Fig. \ref{fig:3DCon}, the sTrICP algorithm is locally convergent for the registration of 3D point sets. As mentioned before, the sTrICP algorithm shares the similar framework with other competing approaches. Therefore, they have almost the same convergent speed. Besides, both the sTrICP and BiTrICP algorithms can always obtain good scaling registration for other range image pairs. But the BiTrICP algorithm is required to establish bidirectional correspondences for each point in the data shape, it costs more time than the sTrICP algorithm, which only needs to establish the forward correspondence for each point in the data shape.

 For the range image with high overlapping percentage, such as the Stanford Happy Buddha, the PCA method can provide good estimation of the scaler factor. Therefore, it is easy for the TrICPBs algorithm to obtain accurate results for scaling registration. For the range image with common overlapping percentages, such as the Stanford Bunny and Dragon, the PCA method is unable to provide good estimation of the scaler factor. Subsequently, a narrow range of boundaries may not cover the real scale factor, so the TrICPBs algorithm can never obtain the accurate results for scaling registration of range image pairs. Otherwise, a wide range of boundaries will make the TrICPBs algorithm easy to trap into local minimum around the low boundary. Therefore, the performance of the TrICPBs algorithm is more worse than that of other two approaches.

Therefore, the proposed approach can achieve the scaling registration of range images with the best performance among these competing approaches.

\section{Application: Grid map merging}
Mapping has been considered to be an important prerequisite of truly autonomous robots. Its efficiency can potentially be  increased in a significant way by utilizing multiple robots exploring different parts of the environment. Therefore, a key question is how to merge maps built by the different robots into a single global map.

\begin{figure}[htp]
\begin{center}
\subfigure[]{\label{fig:plot-a}\includegraphics[scale=1.0]{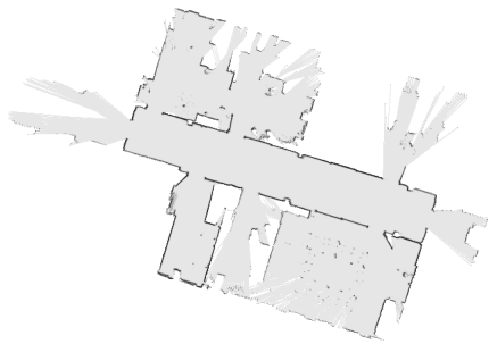}}
\subfigure[]{\label{fig:plot-b}\includegraphics[scale=0.9]{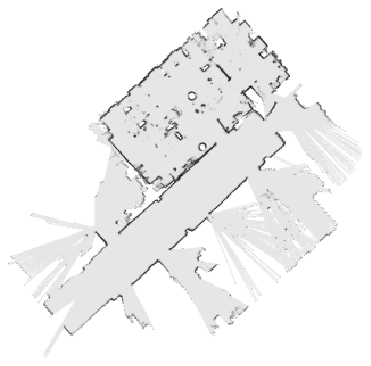}}
\subfigure[]{\label{fig:plot-c}\includegraphics[scale=1.0]{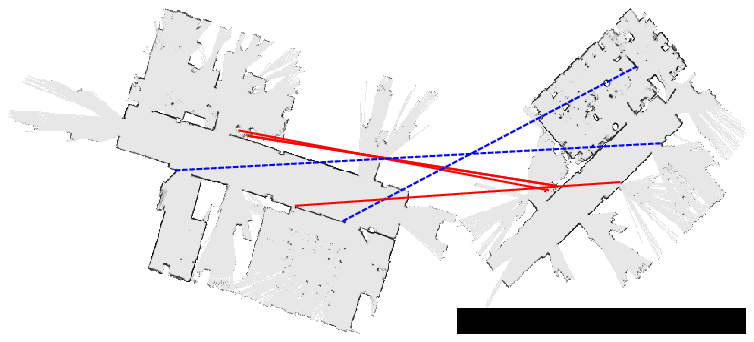}}
\subfigure[]{\label{fig:plot-d}\includegraphics[scale=0.5]{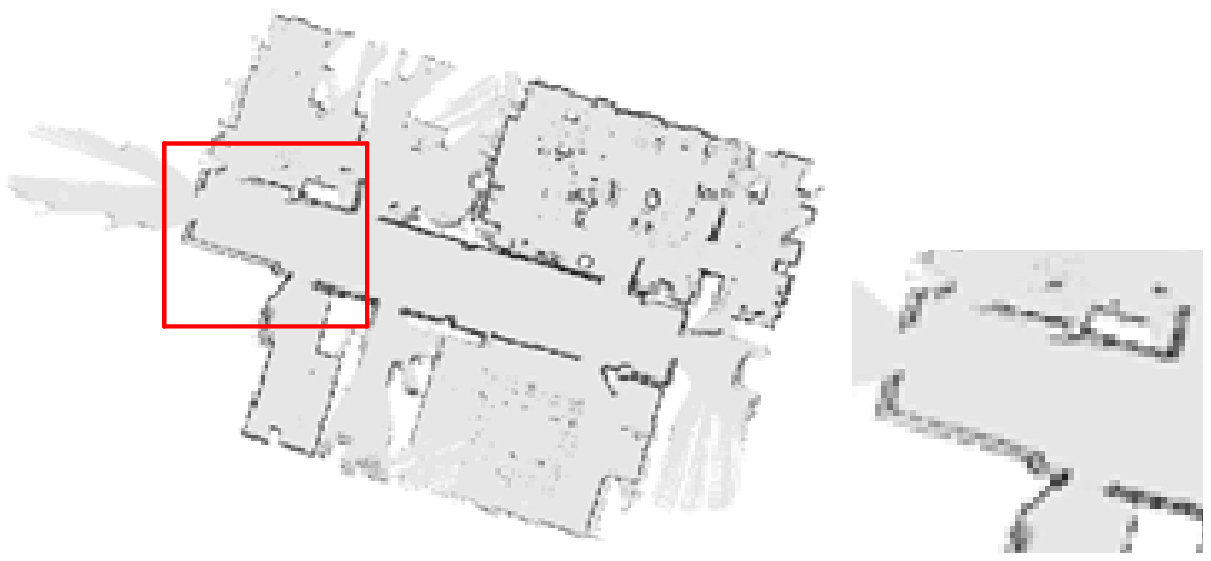}}
\subfigure[]{\label{fig:plot-e}\includegraphics[scale=0.5]{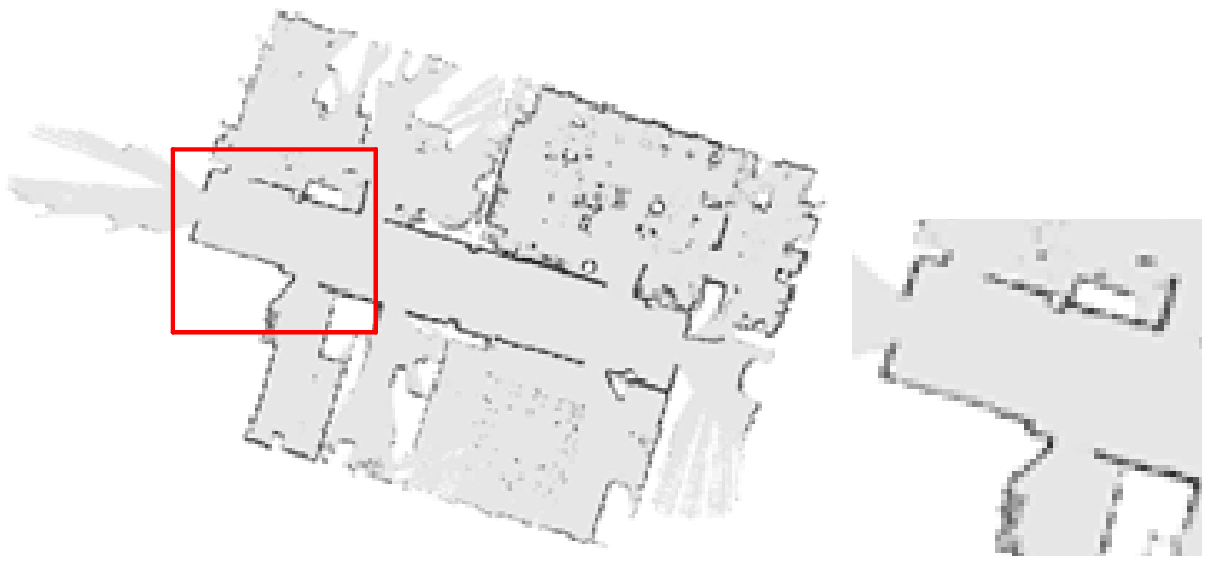}}
\end{center}
\caption{ The demonstration of merging grid map. (a) Reference map (4cm). (b) Non-reference map (5cm). (c) SIFT features matched between maps to be merged, where solid red lines denote the geometrically
consistent matches and dashed blue lines indicate the inconsistent matches. (d) Coarse merged result based on the initial parameters. (c) Fine merged result based on the sTrICP algorithm.}
\label{fig:Gridmap}
\end{figure}

To verify the proposed approach, an experiment was performed on a real robot dataset, which is available for public access \cite{Robot}. As shown in Fig. \ref{fig:Gridmap}, there are two grid maps with different resolutions, which were produced by the robot exploring different parts of the same environment. To achieve grid maps, the scale-invariant feature transform (SIFT) features \cite{Lowe04} were extracted and matched for two grid maps to be merged. Then, the random sample consensus (RANSAC) algorithm \cite{Lowe03} was applied to confirm two geometrically consistent feature matches, which can be utilized to estimate the initial transformation for the scaling registration algorithm. As shown in Fig. \ref{fig:plot-c}, the initial transformation is inaccurate and the merging map is coarse. Subsequently, the edge points can be extracted from these two grid maps. According to our previous work \cite{Ma16}, the problem of merging grid maps with different resolutions can be reasonably viewed as the scaling registration of the partially overlapping point sets. Therefore, the sTrICP algorithm can be applied to the registration of extracted edge point sets and obtain the accurate scaling transformation. Finally, these two grid maps can be merged into a single global map, which is more fine than the initial merged one.

\begin{figure}[htp]
\begin{center}
\subfigure[]{\label{fig:Lplot-a}\includegraphics[scale=0.7]{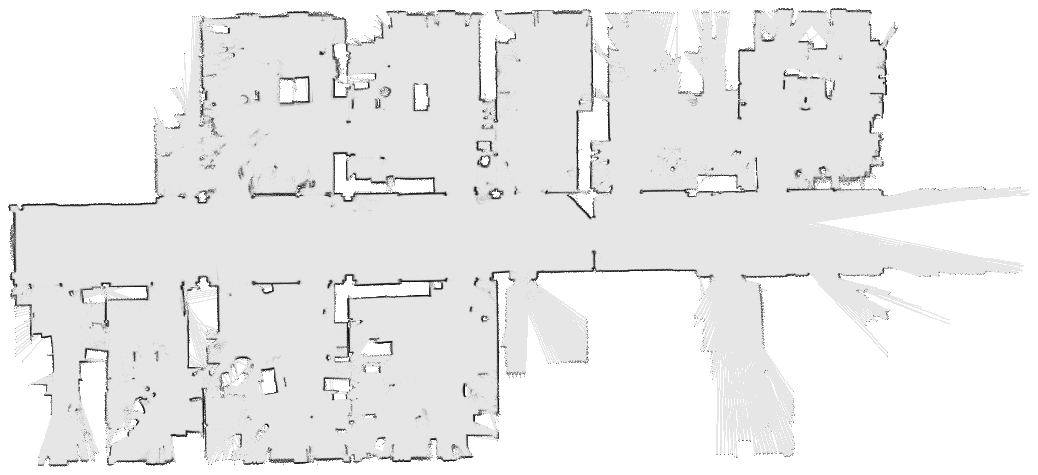}}
\subfigure[]{\label{fig:Lplot-b}\includegraphics[scale=0.7]{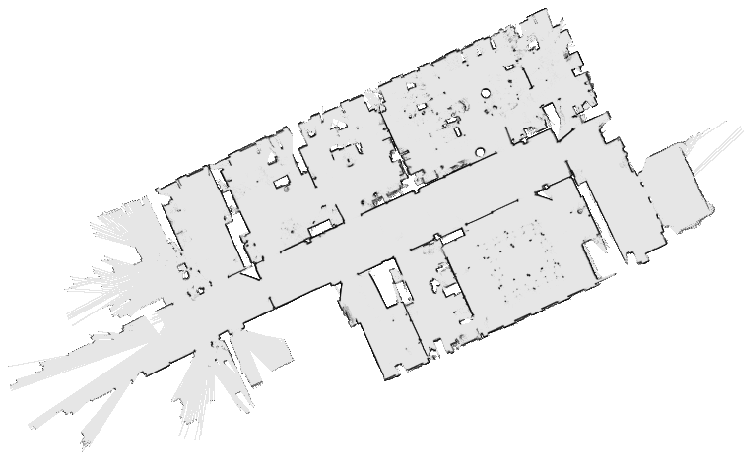}}
\subfigure[]{\label{fig:Lplot-c}\includegraphics[scale=0.7]{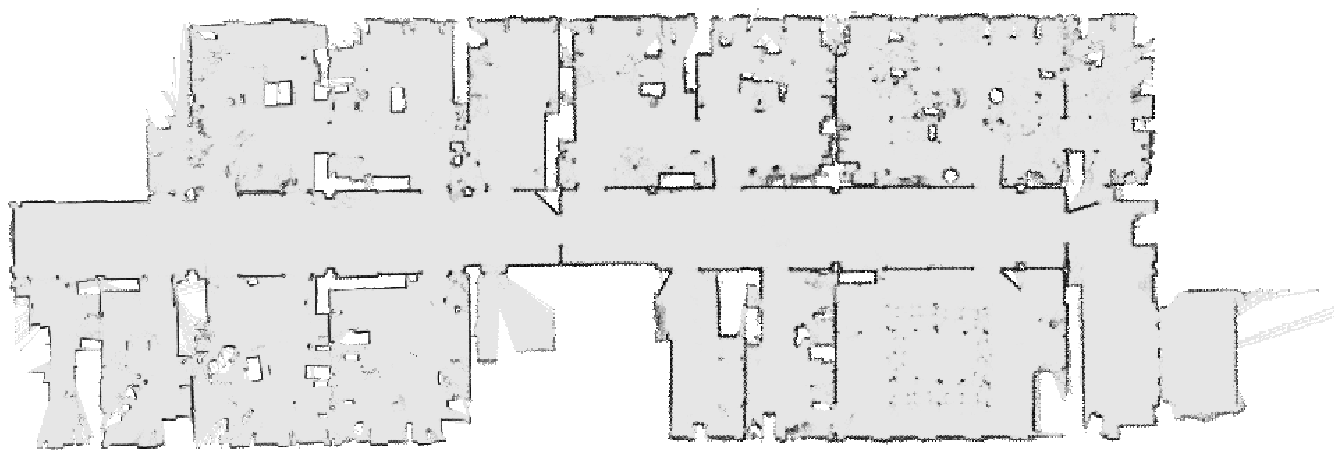}}
\end{center}
\caption{ The application of the proposed approach. (a) Reference map (3cm). (b) Non-reference map (5cm). (c) The merged result based on the sTrICP algorithm (3cm).}
\label{fig:LGridmap}
\end{figure}

To further illustrate the application, Fig. \ref{fig:LGridmap} also displays another grid map merging result, where the two grid maps to be merged are in different resolutions. After applying the sTrICP algorithm, we can obtain a global grid map, which has the similar resolution as the reference grid map.

\section{Conclusions}
This paper proposes a novel approach for the scaling registration of $m$-D point sets, which are absolutely or partially overlapping to each other. Firstly, for the absolutely overlapping point sets, a novel objective function is designed by imposing the emphasis on the scale factor. Similar to other registration problems, the new objective function also includes the item of the least-square error, which exits as its numerator. The significant difference is the introduction of the denominator item, which is the square of the scale factor and can avoid the ill-posed problem caused by the scale factor. Subsequently, a variant of the ICP algorithm is then proposed to solve this new objective function. What's more, this approach is further extended to align partially overlapping point sets for the practical applications.
Experimental results illustrate that the proposed approach can achieve the scaling
registration of $m$-D point sets with good performance.

\section*{Acknowledgments}
This work is supported by the National Natural Science Foundation of
China under Grant nos. 61573273, 61573280 and 61503300.

\end{document}